# Inference of Fine-grained Attributes of Bengali Corpus for Stylometry Detection

Tanmoy Chakraborty and Sivaji Bandyopadhyay

*Abstract*—Stylometry, the science of inferring characteristics of the author from the characteristics of documents written by that author, is a problem with a long history and belongs to the core task of Text categorization that involves authorship identification, plagiarism detection, forensic investigation, computer security, copyright and estate disputes etc. In this work, we present a strategy for stylometry detection of documents written in Bengali. We adopt a set of fine-grained attribute features with a set of lexical markers for the analysis of the text and use three semi-supervised measures for making decisions. Finally, a majority voting approach has been taken for final classification. The system is fully automatic and language-independent. Evaluation results of our attempt for Bengali author's stylometry detection show reasonably promising accuracy in comparison to the baseline model.

## I. INTRODUCTION

STYLOMETRY is an approach that analyses text in text mining e.g. novels, stories, dramas written by authors, trying to measure the author's style, rhythm of his pen, subjection of his desire, prosody of his mind by choosing some attributes that are consistent throughout his writing and play the linguistic fingerprint of that author. In other words, stylometry is the application of the study of linguistic style, usually with reference to written text that concerns the way of writing rather than its contents. Computational Stylometry is focused on subconscious elements of style less easy to imitate or falsify.

Stylistic analysis that has been done by Croft [2] claimed that for a given author, the habits "*of style*" are not affected "*by passage of time, change of subject matter or literary form. They are thus stable within an author's writing, but they have been found to vary from one author to another*" [8]. However, stylometric authorship attribution can be considered as a typical clustering, classification and association rule problem, where a set of documents with known authorships are used for training and the aim is to automatically determine the corresponding author of an anonymous text, but the way of selecting the appropriate features is not focused in that sense and vary from one research to other.

Most of the authorship identification studies are better at dealing with some closed questions like (i) who wrote this, A or B, (ii) if A wrote these, did he also writes this, (iii) how likely is it that A wrote this etc. The main target in this study is to build a decision making system that enables users to predict and to choose the right author from an anonymous author's novel under consideration, by choosing various lexical, syntactic, analytical features known as *stylistic markers*. The system uses three semi-supervised, reference based measurements (Cosine-similarity, Chi-square measurement and Euclidean distance) which behave as an expert opinion to map the testing documents to the appropriate authors. Without focusing much on the distributional lexical measures like vocabulary richness or frequency of individual word counts, we mainly focus on some low-level measures (sentence count, word count, punctuation count, length of words and sentences etc.), phrase level measures (noun chunk, verb chunk, etc.) and context level measures (number of dialog, length of dialog, sentence structure analysis etc.). Additionally, we propose a baseline system for Bengali stylometry analysis using *vocabulary richness function*. The present attempt basically deals with the microscopic observation for the stylistic behaviours of the articles written by the famous novel laureate Rabindranath Tagore long years back and tries to disambiguate them from the anonymous articles written by some other authors in that period.

The paper is organized as follows. In Section 2, related researches on stylometry and authorship identification in other language and their approaches are described. In Section 3, our approach is detailed along with the extracted features and classification models used in this experiment. The experimental results compared to the baseline system are described in Section 4. The experimental results are analysed in Section 5 and the conclusions are drawn in Section 6.

## II. RELATED WORKS

Stylometry, which may be considered as an investigation of "Who was behind the keyboard when the document was produced?" or "Did Mr. X wrote the document or not?" is a long term study mainly in forensic investigation department that started from late Nineties. In the past, where stylometry emphasized the rarest or most striking elements of a text, contemporary techniques can isolate identifying patterns even in common parts of speech. The pioneering study on authorship attributes identification using word-length histograms appeared at the very end of nineteen century [6]. After that, a number of studies based on content analysis [5], computational stylistic approach [4], exponential gradient learn algorithm [7], Winnow regularized algorithm [9], Support Vector Machine based approach [3] etc. have been proposed for various languages like English and Portuguese. Recently, research has started to focus on authorship attribution on larger sets of authors: 8 [11], 20 [12], 114 [13], or up to thousands of authors [14]. The use of computers regarding the extraction of stylometrics has been limited to auxiliary tools (i.e. simple program for counting user-defined features fast and reliably). Hence, authorship attribution studies so far may be looked like *computer-assisted*, not *compute-based*. As a beginning of Indian language stylometry analysis, our research does not consider any manual intervention for extracting features (like identification of some high frequent start-up words), moreover we have dealt with a number of large-size non-homogeneous texts since they are composed of dialogues, narrative parts etc. and try to build a language and text-length independent system for attribute analysis.

### III. OUR APPROACH

The methodology used in this work generally depends on the combination of 76 fine-grained style-markers for feature engineering and three semi-supervised approaches for decision making. As an initial attempt, we have decided to work with the simple approach like statistical measurement, analyze the drawbacks and further go beyond for working with other machine learning or hybrid approaches. Furthermore, the reasons for not attempting with the methods described in the related work section are as follows: the content analysis is one of the earliest types of computations, also for exponential and Winnow algorithms as both are purely mathematical models and the SVM based method has a strong affinity to the language for which the system is designed. Currently, authorship attribute studies are dominated by the use of lexical measures. In a review paper [1], the author asserted that:

" ..... yet, to date, no stylometrist has managed to establish a methodology which is better able to capture the style of a text than based on lexical items."

For this reason, in order to set a baseline for the evaluation of the proposed method, we have decided to implement a lexical based approach called *vocabulary richness*. Detailed discussion about the baseline system and our approach are mentioned in Section 4.

#### A. Proposed Methodology Design

As mentioned, the proposed stylistic markers used in this study take full advantage of the analysis of the distributed contextual clues as well as full analysis by natural language processing tools. The system architecture of the proposed stylometry detection system is shown in Figure 1. In this section, we first describe brief properties of different components of the system architecture and then the set of stylistic features is analytically presented. Finally the classification methods are elaborated with brief description of their functionalities.

*1) Textual Analysis*

Basic pre-processing before actual textual analysis has been done so that stylistic markers are clearly viewed for further analysis by the system. Token-level markers discussed in the next Section, are extracted from the pre-processed corpus. Then parsing using Shallow parser[1] has been done to separate the sentence and the chunk boundaries and parts-of-speech. From this parsed text, chunk-level and context-level markers are identified.

*2) Stylistic Features Extraction*

Stylistic features have been proposed as more reliable style markers than for example, word-level features since they are not under the conscious control of the author. To allow the selection of the linguistic features rather than n-gram terms, robust and accurate text analysis tools such as lemmatizers, part-of-speech (POS) taggers, chunkers etc. are needed. We have used the Shallow parser, which gives a parsed output of a raw input corpus. It tokenizes the input, performs a part-of-speech analysis, looks for chunks and inflections and a number of other grammatical relations. The stylistic markers which have been selected in this experiment are coarsely classified into three categories and discussed in the Table I.

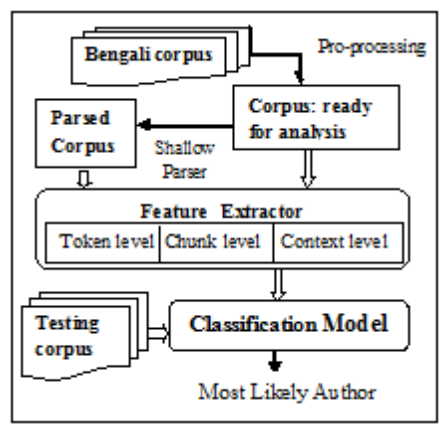

Fig. 1. Architecture of the stylometry detection system.

Sentences are detected using the sentence boundary markers mainly *'dari'* or *'viram'* ('।'), question marks ('?') or exclamation notation ('!') in Bengali. Sentence length and word count are the traditional and well-defined measures in authorship attribute studies and punctuation count is another interesting characteristics of the personal style of a writer. Chunk or phrase level markers are indications of various stylistic aspects, e.g., syntactic complexity, formality etc. Out of all detected chunk sets, mainly nine chunk types have been considered in this experiment. They are noun chunk (NP), verb-finite chunk (VGF), verb-non-finite chunk (VGNF), gerunds (VGNN), adjective chunk (JJP), adverb chunk (RBP), conjunct phrase (CCP), chunk fragment (FRAGP) and others (OTHERS). Shallow parser identifies 25 Part Of Speech (POS) categories. Among them, 24 POSs have been taken into consideration except UNK. Words tagged with UNK are unknown words and are verified by Bengali monolingual dictionary. Since Shallow parser is an automated text-processing tool, the style markers of the above levels are measured approximately. Depending on the complexity of the text, the provided measures may vary from real values which

---
[1] http://ltrc.iiit.ac.in/analyzer/bengali

Inference of Fine-grained Attributes of Bengali Corpus for Stylometry Detectioncan only be measured using manual intervention. Making the system fully automated, the system performance depends on the performance of the parser. As we can see in the Table I that each marker is defined as a percentage measure of the ratio of two relevant measures, this approach was followed in order to work with text-length independent style markers as possible. However, it is worth noting that we do not claim that the proposed set of 76 markers is the final one. It could be possible to split them into more fine-grained measures e.g. F21 can be split into separate measures i.e. individual occurrence of the punctuation symbols (comma per word, colon per word, dari per word etc.). Here, our goal is to make an attempt towards the investigation of Bengali author's writing style and to prove that an appropriately defined set of such style markers performs better than the traditional lexical based approaches.

TABLE I
FINE-GRAINED STYLISTIC FEATURES

| Coarse-grained Classification | Stylistic Markers | Description | Total |
|---|---|---|---|
| Token-level | F1 to F10 | Word length (1 to 9 and above) in % | 10 |
| | F11 to F20 | Words per sentence (0-10, 10-20 and so on, up to 80-90 and above) in % | 10 |
| | F21 | Punctuations per word in % | 1 |
| | F22 to F31 | Individual punctuations in % ( 10 punctuations) | 10 |
| Chunk-level | F32 to F40 | Detected NP, VGF, VGNF, VGNN, JJP, RBP, CCP, FRAGP, OTHER out of total chunks in % | 9 |
| | F41 to F49 | Average words included in all above mentioned chunks in % | 9 |
| | F50 to F73 | Individual percentage of detected POS (24) by Shallow parser | 24 |
| Context-level | F74 | Average words per dialog in % | 1 |
| | F75 | Words not included in the dictionary including Named-Entities in % | 1 |
| | F76 | Hapax-legomena count out of all words in % | 1 |

*3) Classification Model*

A number of discriminative models based on statistical and machine learning measures, such as Bayesian Network, decision trees, neural networks, support vector machines, K-nearest neighbour approach etc. are available for text categorization. In this experiment, three semi-supervised, reference-based classification models have been used: (1) Cosine-similarity measurement, (2) Chi-square measure and (3) Euclidean distance. These are briefly discussed below.

***Cosine-similarity measurement:*** Cosine-similarity is a measure of similarity between two vectors of ***n*** dimensions by finding the cosine of the angle between them, often used to compare documents in text mining. Given two vectors of attributes, ***R*** and ***T***, the cosine similarity, ***θ*** is represented using a dot product and magnitude as:

$$Similarity = \cos(\theta) = \frac{R.T}{|R|.|T|} = \frac{\sum_{i=1}^{n} r_i . t_i}{\sqrt{\sum_{i=1}^{n} r_i^2} * \sqrt{\sum_{i=1}^{n} t_i^2}}$$

The resulting similarity ranges from −1 meaning exactly opposite, to 1 meaning exactly the same, with 0 usually indicating independence, and in-between values indicating intermediate similarity or dissimilarity. Here, n is the number of features (i.e., 76) that act as dimensions of the vectors and $r_i$ and $t_i$ are the features of reference and test vectors respectively.

***Chi-square measure***: Chi-square is a statistical test commonly used to compare observed data with the expected data according to a specific hypothesis. That is, chi-square ($\chi^2$) is the sum of the squared differences between observed (***O***) and the expected (***E***) data (or the deviation, ***d***), divided by the expected data in all possible categories.

$$\chi^2 = \sum_{i=1}^{n} \frac{(O_i - E_i)^2}{E_i}$$

Here, the mean of each cluster is used as the observation data for that cluster and used as reference ***O***. ***n*** is the number of features and $O_i$ is the observed value of the i[th] feature. The Chi Square test gives a value of $\chi^2$ that can be converted to Chi Square ($c^2$) using chi-square table which is an n×n matrix with row representing the degree of freedom (i.e., difference between the number of rows and columns of the contingency matrix) and column representing the probability we expect. This can be used to determine whether there is a significant difference from the null hypothesis or whether the results support the null hypothesis. After comparing the chi-squared value in the cell with our calculated $\chi^2$ value, if the $\chi^2$ value is greater than the 0.05, 0.01 or 0.001 column, then the goodness-of-fit null hypothesis can be rejected, otherwise accepted.

***Euclidean distance:*** The Euclidean distance between two points, ***p*** and ***q*** is the length of the line segment. In Cartesian coordinates, if ***p*** = (*p₁, p₂... pₙ*) and ***q*** = (*q₁, q₂,..., qₙ*) are two points in Euclidean n-space, then the distance from ***p*** to ***q*** is given by:

$$d(p,q) = \sqrt{\sum_{i=1}^{n}(p_i - q_i)^2}$$

where, ***n*** is the number of features or dimension of a point, ***p*** is the reference point (i.e. mean vector) of each cluster and ***q*** is the testing vector. For every test vector, three distances from three reference points have been calculated and smallest distance defines the probable cluster.

81	*Polibits (44) 2011*

## IV. EXPERIMENT

### A. Corpus

Resource acquisition is one of the most important challenges to work with resource constrained languages like Bengali. The system has used thirty stories in Bengali written by the noted Indian Nobel laureate Rabindranath Tagore[2]. Among them, we have selected twenty stories for training purpose and rest for testing. We choose this domain for two reasons: firstly, in such writings the idiosyncratic style of the author is not likely to be overshadowed by the characteristics of the corresponding text genre; secondly, an earlier work [10] has worked on the corpus of Rabindranath Tagore to explore some of the stylistic behaviours of his writing. To differentiate them from other author's articles, we have selected 30 articles from author A and 30 articles of other authors[3]. In this way, we have three clustered set of documents identified as articles of Author R (Tagore's articles), Author A and others (O). This paper focuses on two topics: (i) the effort of earlier works on feature selection and learning and (ii) the effort of limited data in authorship detection.

### B. Baseline System

In order to set up a baseline system, we proposed traditional lexical based methodology called *vocabulary richness*. Among the various measures like Yule's K measure, Honore's R measure, we have taken most typical one as the type-token ratio *(V/N)* where *V* is the size of the vocabulary of the sample text and *N* is the number of tokens which forms the simple text. We have gathered dimensional features of the articles of each cluster and averaged them to make a mean vector for every cluster.

TABLE II
CONFUSION MATRIX OF THE BASELINE SYSTEM

|   | R | A | O | e (Error) |
|---|---|---|---|---|
| R | 6 | 0 | 4 | 0.40 |
| A | 7 | 2 | 1 | 0.80 |
| O | 5 | 2 | 3 | 0.70 |
| Average error | | | | 0.63 |

So these three mean vectors indicate the references of three clusters respectively. Now, for every testing document, similar features have been extracted and a test vector has been developed. Now, using Nearest-neighbour algorithm, we have tried to identify the author of the test documents. The results of the baseline system are shown using confusion matrix in Table II. Each row shows classification of the ten texts of the corresponding authors. The diagonal contains the correct classification. The baseline system achieves 37% average accuracy. Approximately 60% of average accuracy error (for author A and O) is due to the wrong identification of the author as Author R.

### C. Performance of Our System

We have discussed earlier that our classification model is based on three statistical techniques. A voting approach combining the decision of the three models for each test document have also been measured for expecting better results.

The confusion matrix in Table III and IV shows that Chi-square measure has relatively less error (46%) rate compared to other measures. A majority voting technique has an accuracy rate of 63% which is relatively better than others. In the case when the three statistical techniques produce different results, the result of Chi-square measure has been taken as correct result because it has given more accuracy compared to the others when measured individually.

TABLE III
CONFUSION MATRIX OF OUR SYSTEM (PART I)

| | Cosine-similarity | | | | Chi-square measure | | | |
|---|---|---|---|---|---|---|---|---|
| | R | A | O | e | R | A | O | e |
| R | 5 | 2 | 3 | 0.5 | 7 | 3 | 0 | 0.3 |
| A | 3 | 6 | 1 | 0.4 | 5 | 4 | 1 | 0.6 |
| O | 4 | 1 | 5 | 0.5 | 4 | 1 | 5 | 0.5 |
| Average error | | | 0.46 | Average error | | | 0.46 |

TABLE IV
CONFUSION MATRIX OF OUR SYSTEM (PART II)

| | Euclidean distance | | | | Combined voting | | | |
|---|---|---|---|---|---|---|---|---|
| | R | A | O | e | R | A | O | e |
| R | 6 | 2 | 2 | 0.4 | 8 | 2 | 0 | 0.2 |
| A | 4 | 4 | 2 | 0.6 | 4 | 5 | 1 | 0.5 |
| O | 3 | 2 | 5 | 0.5 | 2 | 2 | 6 | 0.4 |
| Average error | | | 0.5 | Average error | | | 0.37 |

## V. DISCUSSION

Form the experimental results, it is clear that statistical approaches show nearly similar performance and accuracies of all of them are around 50%. Also the major sources of the errors are for the inappropriate identification of author as Author R. From the figure 2, we can see that the system looks little bit biased towards the identification of Rabindranath Tagore as author of the test documents. In all cases, the bar graphs for Author R are higher than others. The reason behind this is the acquisition of resources. Developing appropriate corpus for this study is itself a separate research area and takes huge amount of time. Furthermore, the collected articles from other authors are heterogeneous and not domain constrained.

Our studies in future will be planned to focus on the identification of the unpublished articles of Rabindranath Tagore. For this, more microscopic observation in various fields of his writings will be needed. Here we only try our experiments on the stories of the writer. The success of the system lies not on the correct mapping of the articles to their corresponding three authors but to filter all the inventions of

---

[2] http://www.rabindra-rachanabali.nltr.org
[3] http://banglalibrary.evergreenbangla.com

Rabindranath Tagore from a bag of documents and the more the accuracy of the filtering, the more the accuracy of the system. Apart from being the first work of its kind for Bengali language, the contributions of this experiment can be identified as: (i) application of statistical approach in the field of stylometry, (ii) development of classification algorithm in n-dimensional vector space, (iii) developing a baseline system in this field and (iv) more importantly, working with the great writings of Rabindranath Tagore to reveal his swinging of thought and dexterity of pen when writing articles.

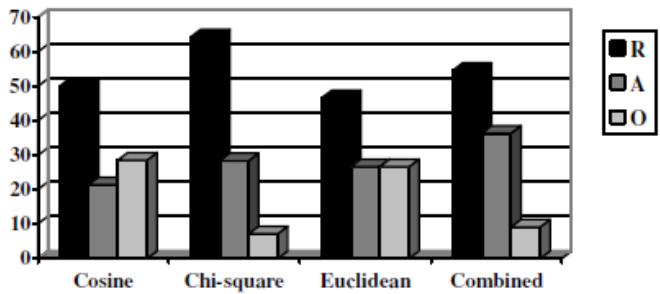

Fig. 2. Error analysis of different approaches.

## VI. CONCLUSIONS

This paper introduced the use of a large number of fine-grained features for stylometry detection. The presented methodology can also be used in author verification task i.e. the verification of the hypothesis whether or not a given person is the author of the text under study. The methodology can be adopted for other languages since maximum of the features are language independent. The classification is very fast since it is based on the calculation of some simple statistical measurements. Particularly, it appears from our experiments that texts with less word are less likely to be classified correctly. For that, our system is little biased towards the stylometry of Rabindranath Tagore. It is due to the lack of the large number of resources of other authors under study. However from this preliminary study, future works are planned to increase the database with more fine-grained features and to identify more context dependent attributes for further improvement.